\documentclass[times, review, 10pt]{elsarticle}
\usepackage{amssymb}
\usepackage{amsthm}
\usepackage[figuresright]{rotating}
\usepackage{amsmath}
\usepackage{mathrsfs}
\usepackage{newtxmath}
\usepackage{color, colortbl}
\usepackage{amsfonts}
\usepackage{mathtools}
\usepackage{algorithm}
\usepackage{algorithmic}
\usepackage{soul}
\usepackage{url}
\usepackage[utf8]{inputenc}
\usepackage{caption}
\usepackage{subcaption}
\usepackage{graphicx}
\usepackage{booktabs}
\usepackage{xcolor}
\usepackage{multirow}
\usepackage{verbatim}
\usepackage{float}
\usepackage{textcomp}
\usepackage{lipsum}
\usepackage{stfloats}
\usepackage{array}
\usepackage{comment}
\usepackage{makecell}

\usepackage{setspace} 

\usepackage{hyperref}
\hypersetup{
    colorlinks=true,
    linkcolor=blue,
}
\journal{Pattern Recognition}
\begin{document}

\begin{frontmatter}

\title{StreamEMS: Streaming Video Understanding with Self-Evolving Memory Scheme for Vision-Language Models}

\author[inst1]{Yuxin Liu}
\ead{yx.liu5@siat.ac.cn}
\author[inst2]{Peiqin Zhuang\corref{cor1}}
% \ead{pzhu4617@uni.sydney.edu.au}
\ead{zpq0316@163.com}
\author[inst1]{Yali Wang\corref{cor1}}
\cortext[cor1]{Equal corresponding authors}
\ead{yl.wang@siat.ac.cn}

\affiliation[inst1]{organization={Shenzhen Institute of
Advanced Technology, Chinese Academy of Sciences},
            city={Shenzhen}, country={China}}
\affiliation[inst2]{organization={Shanghai AI Laboratory},
            city={Shanghai}, country={China}}

\begin{abstract}
Recently, many streaming video understanding methods have been proposed by constructing an external memory to store historical data for computational reduction. Most methods focus on optimizing the injection procedure of current data (write) and retrieving informative historical data (read) from memory, while overlooking the opportunity to further enhancing the representational capability of memory itself.
In this work, we present \textbf{StreamEMS}, a general mechanism for improving streaming video understanding by re-structuring the historical data stored in memory through self-evolving memory scheme, enabling more informative and robust memory representations.
Specifically,
we first introduce a Semantic Evolution Module to evolve the memory into more information-dense representations by exploiting informative memory entities discovered via progressively shrinking semantic scales from coarse to fine.
In addition, we further introduce a Prior-informed Evolution Module to evolve memory into more robust representations by leveraging prior memory distributions to refine the current memory state.
We validate the effectiveness of our proposed designs on widely-used streaming video understanding datasets, i.e., OVO-Bench and StreamingBench, and the results showcase that our method performs better than other methods. 
Moreover,  the advantage of our method becomes consistently evident even under high token usage drop rate settings, 
indicating the effectiveness and robustness
of our method in unleashing the potential of the memory itself.
\end{abstract}


\end{frontmatter}

\section{Introduction}
\label{sec:intro}

Recently, research on streaming video understanding has become prominent, driven by the growing demand from streaming video platforms. Different from offline video understanding~\cite{,WANG2020107248longvideovqa-matching-attention-model,YUAN2026113740coffee-mate-keyframe-selection-vqa,HE2026114009scan-focus-amplify-guidance-aware-answer-vqa}, the streaming scenario requires the model to answer questions or even make predictions without access to future information, which demands a deeper understanding of long-duration past visual context~\cite{chen2024videollmonline,zhang2025flash,zhang2025infinite}.

Generally, many approaches have been proposed by explicitly constructing an external memory~\cite{zhang2025flash,huang2025onlineOVBench} or using memory as an implicit approach~\cite{yao2025timechat,qian2025dispider,ding2025streammind,wang2025streambridge} to cope with the problem. Accordingly, they design various read/write\footnote{Write denotes injecting current data into memory while read means retrieving historical data from memory} operations to encapsulate lengthy visual context into memory for later understanding.
For the memory-write paradigm, many methods propose to encode essential information from the current frame into memory via the token dropping~\cite{yao2025timechat} or token merging strategies~\cite{huang2025onlineOVBench,wang2025streambridge,zhang2025flash}.
Alternatively, another line of research leans toward proposing effective mechanisms to enhance reading efficiency for streaming video input scenarios
such as Key-Value Caches~\cite{xu2025streamingvlm,chen2025streamkv,chen2025streamingtom}  etc.

Although those methods have made significant progress on boosting the effectiveness of memory, they mainly focus on memory operations from an external perspective, i.e., read/write. In this case, they overlook the opportunity to further exploit the broader capabilities of memory itself from an internal perspective. A factor would enable the model to discover additional informative and discriminative cues beyond conventional operations, offering strong complementarity and generalization to prior approaches. Similar to this scenario, cognitive neuroscience theory also suggests that the information processing system within human memory does not merely store information but continuously reorganizes previously acquired knowledge internally to form richer representations~\cite{nader2009singleMemoryReconsolidation}.
To discover more distinctive information, the human brain often activates multiple pathways that repeatedly interpret the same information at hierarchical semantic scales, ranging from loose to more rigorous levels, to gradually unveil more salient and meaningful patterns~\cite{jeon2014hierarchicalProcessingCognitive}.
Besides, the brain also exploits leveraging previously retained knowledge as priors to facilitate a deeper understanding of new incoming information~\cite{knill2004bayesianBrain}.

In light of this, we propose to boost native memory in streaming video understanding to mine more distinctive cues for subsequent reasoning.
We achieve this by re-structuring the historical data stored inside memory in a self-evolving way. 
To be specific, we introduce a Semantic Evolution Module (SEM) that constructs semantic graphs to propagate information within memory, and gradually evolves the memory into more information-dense representations by exploiting semantic entities. Those entities are discovered via progressively shrinking semantic scales from coarse to fine.       
On top of that, we further introduce a Prior-informed Evolution Module (PEM) that adopts the Exponential Moving Average (EMA) strategy to integrate prior distribution cues from historical memory into the current memory, fostering discriminative semantic refinement. This factor balances the importance distribution of each memory entry at both the current moment and over history, alleviating the possible bias from concentrating only on the distribution at the current moment.

We validate the effectiveness of our proposed designs on widely-used datasets, i.e., OVO-Bench~\cite{niu2025ovo} and StreamingBench~\cite{lin2024streamingbench}, and the results showcase that our method performs better than other methods. 
Moreover, the advantage of our method becomes consistently evident even under high token usage drop rates,
indicating the effectiveness and robustness of our method in unleashing the potential of the memory itself.
To sum up, our contributions are summarized as follows: \vspace{-0.3cm}

\begin{itemize}
    \item We propose a Semantic Evolution Module (SEM) to evolve the memory into more information-dense representations by leveraging informative memory entities, enhancing the capacity of the existing memory. \vspace{-0.3cm}
    \item We propose a Prior-Informed Evolution Module (PEM) to evolve the memory into more robust representations by leveraging prior memory distributions to refine the current memory state. 
    \vspace{-0.3cm}
    \item We validate the effectiveness of our method, and our method outperforms other approaches on widely-used datasets.
    \vspace{-0.3cm}
\end{itemize}

\section{Related Work}
\label{sec:relwork}

\subsection{Streaming Video Understanding}

Streaming video understanding requires models to process continuous, and possibly infinite video streams~\cite{zhang2025infinite,chen2024videollmonline,li2025lion}.
Typically, the majority of efforts~\cite{zhang2025flash,huang2025onlineOVBench,yao2025timechat,qian2025dispider,ding2025streammind,wang2025streambridge} are devoted to explicitly constructing an external memory to store historical data and compress past visual context for subsequent reasoning.  Meanwhile, they propose various read/write operations that accompany the construction of external memory.
For the memory-write paradigm, many methods adopt token dropping~\cite{yao2025timechat} or token merging~\cite{zhang2025flash,huang2025onlineOVBench,wang2025streambridge,zeng2025streamforest} strategies to encode essential information from the current frame into memory, based on visual similarity. 
Alternatively, a series of approaches~\cite{xu2025streamingvlm,chen2025streamkv,chen2025streamingtom}
have been proposed to improve the memory-read operation, especially in increasing the efficiency of reading out historical tokens. A shared strategy across these approaches is to utilize KV-Cache to decrease the latency of reading tokens of long context.
Different from prior works that center on external memory interactions (e.g., read/write), we aim to explore the intrinsic capacity of memory itself and boost memory internally in a self-evolving fashion. Our method is complementary to existing external-memory techniques and exhibits strong generalization ability across previous methods.

\subsection{Self-Evolving Schemes}
Investigation into self-evolving schemes has been a long-standing research topic.
In machine learning, methods like bootstrap or boosting leverage internal information within data to enhance the model's capacity or data's representation~\cite{efron1994introductionBootstrapping}. In deep learning, various methods follow a similar spirit.
For example, label propagation~\cite{zheng2021online,liulearning} or message passing~\cite{liu2023neural} could be interpreted as a similar strategy in few-shot /semi-supervised learning, where useful information is transferred to unseen/unlabeled data via the affinity graph, making the entire dataset more informative.
Instead of leveraging side information from the graph at the current time, Mean Teacher~\cite{tarvainen2017mean,ge2021self} and Temporal Ensembling~\cite{laine2017temporal} advocate updating the current prediction with previous predictions, via using temporal priors to enhance the robustness and consistency of the prediction with an Exponential Moving Average (EMA).
Effective as these strategies are, the self-evolving paradigm has not been examined in streaming video understanding. In this work, we propose to boost the native memory in a self-evolving manner by leveraging those classical paradigms and formulate them under the streaming video understanding scenario to fully exploit their potential.

\section{Method}
\label{sec:method}

\subsection{Overview}
\label{sec:met-overview}

\begin{figure*}[ht]
    \centering
    \includegraphics[width=\linewidth]{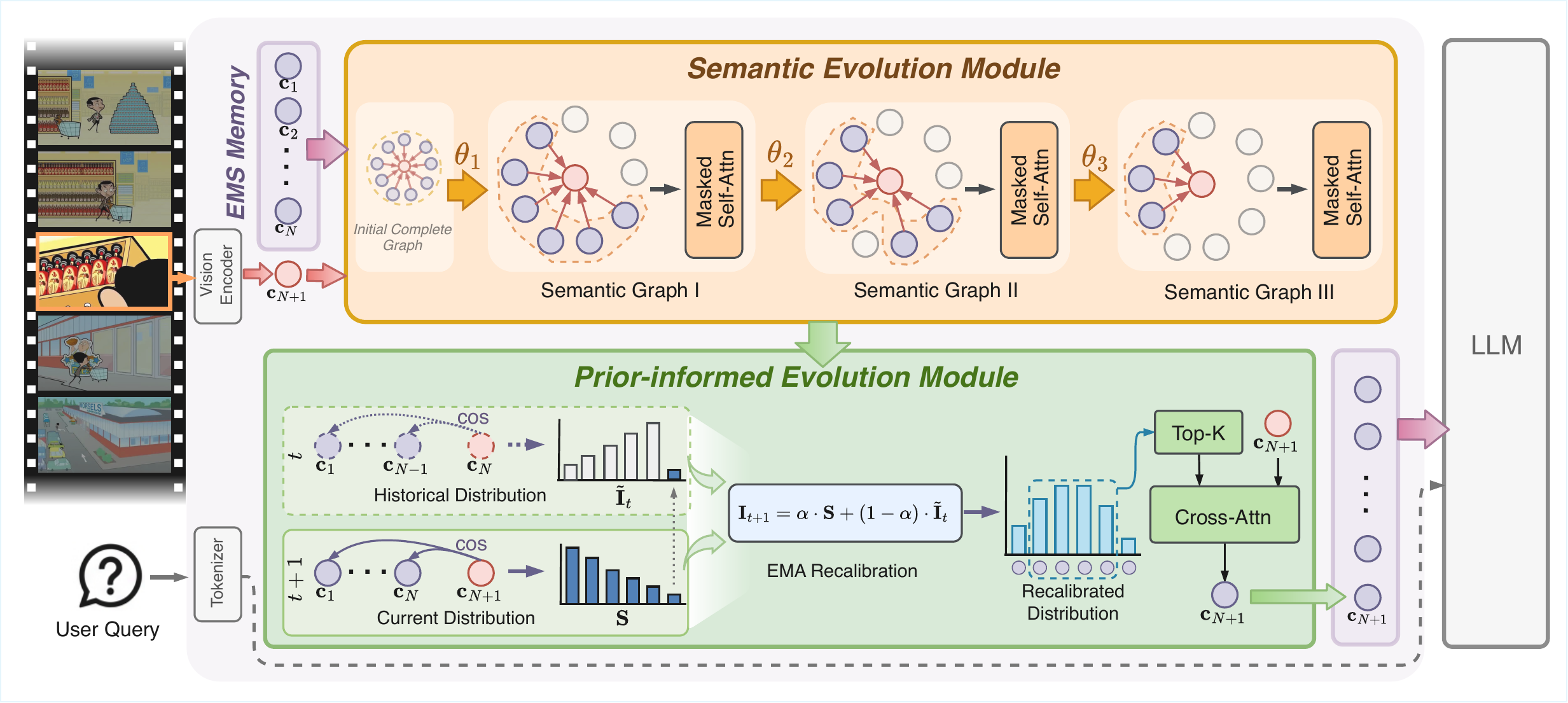}
    \caption{\textbf{An overview of our framework}. We propose to exploit the memory itself with self-evolving memory scheme. 
    In the Semantic Evolution Module, we perform semantic evolution by using a hierarchy of semantic thresholds to select highly related memory entities and aggregate information from them, making the memory more compact.
    After that, we employ the Prior-informed Evolution Module to recalibrate the original similarity distribution using the previous memory distribution in the Exponential Moving Average (EMA) manner.
 }
    \label{fig:framework}
\end{figure*}

As shown in Fig.~\ref{fig:framework}, we follow the widely-used pipeline in streaming video understanding, where a vision encoder and a text encoder are used to extract visual and textual information, respectively. After that, an LLM is used for reasoning over those multi-modal inputs.
Generally, an external memory is constructed and employed to compress lengthy visual contexts for better visual representation.
On this basis, we advocate  further optimizing and evolving the memory in parallel by restructuring the memory entities inside,
with the proposal of the Semantic Evolution Module and Prior-informed Evolution Module.

\subsection{Semantic Evolution Module}
\label{sec:met-sem}

Semantic Evolution Module (SEM) constructs a series of semantic graphs to propagate information within memory, and gradually evolves the memory into more information-dense representations. This can be achieved by exploiting semantic entities within the memory through progressively shrinking semantic scales from coarse to fine.

For the initial construction of memory entities, we represent each of them as a clip-level feature. Given a sequence of video frames, we first use the vision encoder to extract frame-level features. To construct these clip-level features,
we maintain a frame buffer to selectively store frame-level features that exhibit high visual similarity to those already contained within; otherwise, we cease accepting incoming frame-level features for that buffer.  Subsequently, we apply a short-term temporal convolution~\cite{li2020mstcn2} to this set of buffered features to enhance their temporal dependencies, thereby yielding more robust clip-level representations.
Finally, we employ a set of learnable queries to aggregate information from the corresponding clip features and feed these queries into memory as fundamental memory entities.
Those clip-level representations effectively summarize the long-range historical context of the entire video, which significantly reduces computational overhead compared to the traditional frame-by-frame processing paradigm.

\textbf{Progressive Semantic Evolution}. After those clip-level features are placed into memory, we aim to further enrich their semantics as the streaming video input progresses. In practice, existing approaches are limited in achieving this goal. A majority of methods~\cite {yao2025timechat,huang2025onlineOVBench} focus on extending the merging and selective dropping mechanisms across temporally adjacent tokens and incorporating those tokens into memory, while neglecting to further enrich their semantics once the tokens are stored. Thereby, they may limit the potential to further boost the memory's representational power, consequently constraining the model's reasoning capabilities.
Alternatively, other methods adopt a coarse-grained strategy that operates on all memory tokens, e.g., clustering~\cite{zhang2025flash}, to capture information at a large scale. However, such coarse processing may struggle to filter out uninformative signals introduced by irrelevant tokens.

Different from them, we propose to iteratively identify highly relevant memory entities for semantic enhancement, and gradually tighten the similarity threshold $\theta$, to progressively increase the information density of memory.
Specifically, given $N$ memory entities $\{\mathbf{c}_{1}, \mathbf{c}_{2} \dots \mathbf{c}_{N} \}$, we first construct their affinity matrix via measuring the cosine similarity between entities, and use the affinity graph to construct mask matrix $\mathbf{M}$ as follows,

\begin{equation}
    \mathbf{M}_{i,j} = 
    \begin{cases}
    1, & \text{if }  \langle \mathbf{c}_i, \mathbf{c}_j \rangle \geq \theta \\ 
    0, & \textit{otherwise}.
    \end{cases}
\end{equation}
With the mask matrix $\mathbf{M}$ available, we perform masked self-attention among memory entities and leverage it as guidance to propagate distinct semantic information from key/value nodes to the corresponding query node. In addition, we propose to perform semantic evolution in a progressive manner, where the semantic evolution strategy is repeatedly applied with larger similarity thresholds $\theta$.
The incremental increase in $\theta$ allows each node to aggregate information from neighboring entities with more substantial relevance.
This progressive and hierarchical propagation strategy makes memory become more compact and information-dense, which consequently enriches the representational capacity of memory for subsequent reasoning. 

\subsection{Prior-informed Evolution Module}

Beyond the paradigm of propagating information among memory entities at the same time step\footnote{Timestamp denotes the temporal index at the clip granularity, rather than the native frame-level index.},
we further introduce another paradigm that integrates distinct information from the previous memory state to refine the current memory representation. We achieve this by introducing the Prior-informed Evolution Module (PEM), which leverages the Exponential Moving Average (EMA) strategy to integrate historical similarity distributions and thereby refine the current memory state.

Assume that $N$ entities  are stored in the memory at time step $t$.
Based on this, we can derive a similarity distribution $\mathbf{I}_{t} \in \mathbb{R}^{N-1}$ by comparing the most recently stored entity $\mathbf{c}_{N}$ against the remaining $N-1$ entities as follows:
\begin{equation}
    \mathbf{I}_{t} = \{\langle \mathbf{c}_{N}, \mathbf{c}_i \rangle | i = 1, \dots, N-1 \}.
\end{equation}
When obtaining the clip-level feature at time step $t+1$, we integrate it into the memory after it interacts with the previously stored memory entities. We perform the prior-informed evolution strategy during this interaction process. Given the clip-level feature $\mathbf{c}_{N+1}$, we calculate the similarity distribution $\mathbf{S} \in \mathbb{R}^{N}$ between $\mathbf{c}_{N+1}$ and previous $N$ memory entities $\{\mathbf{c}_{1}, \mathbf{c}_{2} \dots \mathbf{c}_{N} \}$ as follows:
\begin{equation}
    \mathbf{S} = \{\langle \mathbf{c}_{N+1}, \mathbf{c}_i \rangle | i = 1, \dots, N \}.
\end{equation}
Then, we adopt the EMA strategy to refine the current similarity distribution with the previous one:
\begin{equation}
    % \mathbf{\hat{S}}
    \mathbf{I}_{t+1} = \alpha \cdot \mathbf{S} + (1 - \alpha) \cdot \tilde{\mathbf{I}}_{t},
    \label{eq:ema}
\end{equation}
where the smoothing coefficient $\alpha$
balances the prior similarity distribution and drives its evolution toward refining the new one.
$\tilde{\mathbf{I}}_{t}$ denotes a variant of $\mathbf{I}_{t}$, where we append the last element of $\mathbf{S}$ to address the dimensional misalignment between $\mathbf{S}$ and $\mathbf{I}_{t}$.

After that, we adopt a Top-K strategy to further select highly informative entities from existing memory entities and use them to
exchange information with the current memory entities via cross-attention.
We validate the effectiveness of this EMA evolution strategy and its hyperparameters in Section~\ref{sec:exp.ablations}.

\section{Experiments}

\subsection{Implementation Details}

The architecture of our method is built on Qwen2.5-VL-7B model~\cite{Qwen2.5-VL}.
We adopt this multimodal model because of its advanced video understanding capacity. For training, we use the instruction dataset used in TimeChat-Online~\cite{yao2025timechat} for supervised fine-tuning, which consists of LLaVA-Video-178K~\cite{zhang2024videoinstructiontuningsyntheticLLaVAVideo178k}, Tarsier2~\cite{yuan2025tarsier2advancinglargevisionlanguage}, VideoChat-Flash~\cite{li2024videochatflash}, and TimeChat-Online-139K~\cite{yao2025timechat}. 
We freeze the vision encoder and only update the parameters of its subsequent modules. We train the model with the AdamW optimizer. The learning rate is  $1\times 10^{-5}$, and the batch size is 128.
For the learning rate adjustment, we use the cosine annealing strategy.  Following standard data processing protocols employed in other works, we sample frames from input videos at the rate of 1 FPS.
Meanwhile,
we set the maximum input resolution of frames to $448\times448$ pixels.
To balance memory capacity with computational efficiency, we set the memory size to 150 and employ a First-In-First-Out (FIFO) queue to manage the streaming memory.

\subsection{Evaluation Benchmarks}
We validate the effectiveness of our method on widely-used online video understanding benchmarks, i.e., OVO-Bench~\cite{niu2025ovo} and StreamingBench~\cite{lin2024streamingbench}. OVO-Bench is a benchmark designed to evaluate video understanding capacity under three online scenarios: Real-time Visual Perception, Backward Tracing, and Forward Active Responding. It consists of 2,800 fine-grained videos with human annotations at the timestamp level. Similarly, StreamingBench is proposed to tackle similar scenarios, like Real-time Visual Understanding, Omni-source Understanding and Contextual Understanding.
It has 900 videos associated with human-annotated QA pairs. For all benchmarks, the accuracy of multiple-choice questions is used as the evaluation metric across different scenarios.

\begin{table*}[t]
\caption{Performance comparison on OVO-Bench, comprising three categories: i) \textit{\textbf{Real-Time} Visual Perception} (OCR: Optical Character Recognition, ACR: Action Recognition, ATR: Attribute Recognition, STU: Spatial Understanding, FPD: Future Prediction, OJR: Object Recognition), ii) \textit{\textbf{Backward} Tracing} (EPM: Episodic Memory, ASI: Action Sequence Identification, HLD: Hallucination Detection), and iii) \textit{\textbf{Forward} Active Responding} (REC: Repetition Event Count, SSR: Sequential Steps Recognition, CRR: Clues Reveal Responding).
}
\label{tab:ovo}
\center
\small
\setlength{\tabcolsep}{1pt}
\resizebox{\textwidth}{!}{
\begin{tabular}{lccccccccccccccccc}
\toprule
\textbf{Model} & \multicolumn{1}{c|}{\textbf{\#Frames}} & \multicolumn{7}{c|}{\textbf{Real-Time}} & \multicolumn{4}{c|}{\textbf{Backward}} & \multicolumn{4}{c|}{\textbf{Forward}} & \textbf{Overall} \\ \cmidrule{3-18} 
\textbf{} & \multicolumn{1}{l|}{\ \ \ \textbf{(FPS)}} & \textbf{OCR} & \textbf{ACR} & \textbf{ATR} & \textbf{STU} & \textbf{FPD} & \textbf{OJR} & \multicolumn{1}{c|}{\textbf{Avg.}} & \textbf{EPM} & \textbf{ASI} & \textbf{HLD} & \multicolumn{1}{c|}{\textbf{Avg.}} & \textbf{REC} & \textbf{SSR} & \textbf{CRR} & \multicolumn{1}{c|}{\textbf{Avg.}} & \textbf{Avg.} \\ \midrule
Human Agents & \multicolumn{1}{c|}{-} & 94.0 & 92.6 & 94.8 & 92.7 & 91.1 & 94.0 & \multicolumn{1}{c|}{93.2} & 92.6 & 93.0 & 91.4 & \multicolumn{1}{c|}{92.3} & 95.5 & 89.7 & 93.6 & \multicolumn{1}{c|}{92.9} & 92.8 \\ \midrule
\multicolumn{18}{c}{\textbf{Closed-source MLLMs}} \\ \midrule
Gemin 1.5 Pro~\cite{team2024gemini15pro} & \multicolumn{1}{c|}{1} & 87.3 & 67.0 & 80.2 & 54.5 & 68.3 & 67.4 & \multicolumn{1}{c|}{70.8} & 68.6 & 75.7 & 52.7 & \multicolumn{1}{c|}{62.3} & 35.5 & 74.2 & 61.7 & \multicolumn{1}{c|}{57.2} & 65.3 \\
GPT-4o~\cite{hurst2024gpt4o} & \multicolumn{1}{c|}{64} & \textbf{69.1} & 65.1 & 65.5 & 50.0 & 68.3 & 63.7 & \multicolumn{1}{c|}{63.6} & 49.8 & 71.0 & 55.4 & \multicolumn{1}{c|}{58.7} & 27.6 & 73.2 & 59.4 & \multicolumn{1}{c|}{53.4} & 58.6 \\ \midrule
\multicolumn{18}{c}{\textbf{Open-source Offline MLLMs}} \\ \midrule
{\color[HTML]{9B9B9B} LLaVA-Next-Video-7B~\cite{zhang2024llavanextvideo}} & \multicolumn{1}{c|}{{\color[HTML]{9B9B9B} 64}} & {\color[HTML]{9B9B9B} 69.8} & {\color[HTML]{9B9B9B} 59.6} & {\color[HTML]{9B9B9B} 66.4} & {\color[HTML]{9B9B9B} 50.6} & {\color[HTML]{9B9B9B} 72.3} & {\color[HTML]{9B9B9B} 61.4} & \multicolumn{1}{c|}{{\color[HTML]{9B9B9B} 63.3}} & {\color[HTML]{9B9B9B} 51.2} & {\color[HTML]{9B9B9B} 64.2} & {\color[HTML]{9B9B9B} 9.7} & \multicolumn{1}{c|}{{\color[HTML]{9B9B9B} 41.7}} & {\color[HTML]{9B9B9B} 34.1} & {\color[HTML]{9B9B9B} 67.6} & {\color[HTML]{9B9B9B} 60.8} & \multicolumn{1}{c|}{{\color[HTML]{9B9B9B} 54.2}} & {\color[HTML]{9B9B9B} 53.1} \\
{\color[HTML]{9B9B9B} LLaVA-OneVision-7B~\cite{lillavaonevision}} & \multicolumn{1}{c|}{{\color[HTML]{9B9B9B} 64}} & {\color[HTML]{9B9B9B} 67.1} & {\color[HTML]{9B9B9B} 58.7} & {\color[HTML]{9B9B9B} 69.8} & {\color[HTML]{9B9B9B} 49.4} & {\color[HTML]{9B9B9B} 71.3} & {\color[HTML]{9B9B9B} 60.3} & \multicolumn{1}{c|}{{\color[HTML]{9B9B9B} 62.8}} & {\color[HTML]{9B9B9B} 52.5} & {\color[HTML]{9B9B9B} 58.8} & {\color[HTML]{9B9B9B} 23.7} & \multicolumn{1}{c|}{{\color[HTML]{9B9B9B} 45.0}} & {\color[HTML]{9B9B9B} 24.8} & {\color[HTML]{9B9B9B} 66.9} & {\color[HTML]{9B9B9B} 60.8} & \multicolumn{1}{c|}{{\color[HTML]{9B9B9B} 50.9}} & {\color[HTML]{9B9B9B} 52.9} \\
{\color[HTML]{9B9B9B} Qwen2-VL-7B~\cite{Qwen2-VL}} & \multicolumn{1}{c|}{{\color[HTML]{9B9B9B} 64}} & {\color[HTML]{9B9B9B} 69.1} & {\color[HTML]{9B9B9B} 53.2} & {\color[HTML]{9B9B9B} 63.8} & {\color[HTML]{9B9B9B} 50.6} & {\color[HTML]{9B9B9B} 66.3} & {\color[HTML]{9B9B9B} 60.9} & \multicolumn{1}{c|}{{\color[HTML]{9B9B9B} 60.7}} & {\color[HTML]{9B9B9B} 44.4} & {\color[HTML]{9B9B9B} 66.9} & {\color[HTML]{9B9B9B} 34.4} & \multicolumn{1}{c|}{{\color[HTML]{9B9B9B} 48.6}} & {\color[HTML]{9B9B9B} 30.1} & {\color[HTML]{9B9B9B} 65.7} & {\color[HTML]{9B9B9B} 50.8} & \multicolumn{1}{c|}{{\color[HTML]{9B9B9B} 48.9}} & {\color[HTML]{9B9B9B} 52.7} \\
{\color[HTML]{9B9B9B} InternVL-V2-7B~\cite{chen2024farInternvl2}} & \multicolumn{1}{c|}{{\color[HTML]{9B9B9B} 64}} & {\color[HTML]{9B9B9B} 68.5} & {\color[HTML]{9B9B9B} 58.7} & {\color[HTML]{9B9B9B} 69.0} & {\color[HTML]{9B9B9B} 44.9} & {\color[HTML]{9B9B9B} 67.3} & {\color[HTML]{9B9B9B} 56.0} & \multicolumn{1}{c|}{{\color[HTML]{9B9B9B} 60.7}} & {\color[HTML]{9B9B9B} 43.1} & {\color[HTML]{9B9B9B} 61.5} & {\color[HTML]{9B9B9B} 27.4} & \multicolumn{1}{c|}{{\color[HTML]{9B9B9B} 44.0}} & {\color[HTML]{9B9B9B} 25.8} & {\color[HTML]{9B9B9B} 57.6} & {\color[HTML]{9B9B9B} 52.9} & \multicolumn{1}{c|}{{\color[HTML]{9B9B9B} 45.4}} & {\color[HTML]{9B9B9B} 50.1} \\
{\color[HTML]{9B9B9B} LongVU-7B~\cite{shen2024longvu}} & \multicolumn{1}{c|}{{\color[HTML]{9B9B9B} 1}} & {\color[HTML]{9B9B9B} 55.7} & {\color[HTML]{9B9B9B} 49.5} & {\color[HTML]{9B9B9B} 59.5} & {\color[HTML]{9B9B9B} 48.3} & {\color[HTML]{9B9B9B} 68.3} & {\color[HTML]{9B9B9B} 63.0} & \multicolumn{1}{c|}{{\color[HTML]{9B9B9B} 57.4}} & {\color[HTML]{9B9B9B} 43.1} & {\color[HTML]{9B9B9B} 66.2} & {\color[HTML]{9B9B9B} 9.1} & \multicolumn{1}{c|}{{\color[HTML]{9B9B9B} 39.5}} & {\color[HTML]{9B9B9B} 16.6} & {\color[HTML]{9B9B9B} 69.0} & {\color[HTML]{9B9B9B} 60.0} & \multicolumn{1}{c|}{{\color[HTML]{9B9B9B} 48.5}} & {\color[HTML]{9B9B9B} 48.5} \\ \midrule
\multicolumn{18}{c}{\textbf{Open-source Online MLLMs}} \\ \midrule
Flash-VStream-7B~\cite{zhang2025flash} & \multicolumn{1}{c|}{1} & 25.5 & 32.1 & 29.3 & 33.7 & 29.7 & 28.8 & \multicolumn{1}{c|}{29.9} & 36.4 & 33.8 & 5.9 & \multicolumn{1}{c|}{25.4} & 5.4 & 67.3 & 60.0 & \multicolumn{1}{c|}{44.2} & 33.2 \\
VideoLLM-online-8B~\cite{chen2024videollmonline} & \multicolumn{1}{c|}{2} & 8.1 & 23.9 & 12.1 & 14.0 & 45.5 & 21.2 & \multicolumn{1}{c|}{20.8} & 22.2 & 18.8 & 12.2 & \multicolumn{1}{c|}{17.7} & - & - & - & \multicolumn{1}{c|}{-} & - \\
Dispider-7B~\cite{qian2025dispider} & \multicolumn{1}{c|}{1} & 57.7 & 49.5 & 62.1 & 44.9 & 61.4 & 51.6 & \multicolumn{1}{c|}{54.5} & 48.5 & 55.4 & 4.3 & \multicolumn{1}{c|}{36.1} & 18.0 & 37.4 & 48.8 & \multicolumn{1}{c|}{34.7} & 41.8 \\
StreamAgent-7B~\cite{yang2025streamagent} & \multicolumn{1}{c|}{1} & 71.2 & 53.2 & 63.6 & 53.9 & 67.3 & 58.7 & \multicolumn{1}{c|}{61.3} & 54.8 & 58.1 & 25.8 & \multicolumn{1}{c|}{41.7} & 35.9 & 48.4 & 52.0 & \multicolumn{1}{c|}{45.4} & {{49.4}} \\
TimeChat-Online-7B~\cite{yao2025timechat} & \multicolumn{1}{c|}{1
} & 74.5 & 48.6 & 68.1 & 48.3 & 69.3 & 59.8 & \multicolumn{1}{c|}{61.4} & 56.9 & 64.9 & 11.8 & \multicolumn{1}{c|}{44.5} & 31.8 & 38.5 & 40 & \multicolumn{1}{c|}{36.8} & 47.6 \\ \midrule
StreamEMS-7B (Ours) & \multicolumn{1}{c|}{1} & 77.2 & 55.1 & 70.7 & 55.1 & 73.3 & 63.6 & \multicolumn{1}{c|}{65.8} & 55.6 & 67.6 & 21.5 & \multicolumn{1}{c|}{48.2} & 33.3 & 40.3 & 47.1 & \multicolumn{1}{c|}{40.2} & \textbf{51.4} \\ \bottomrule
\end{tabular}
}
\end{table*}

\begin{table*}[t]
\caption{Performance comparison on StreamingBench, focusing on \textit{Real-Time Visual Understanding} tasks. Real-Time Visual Understanding encompasses Object Perception (OP), Causal Reasoning (CR), Clips Summarization (CS), Attribute Perception (ATP), Event Understanding (EU), Text-Rich Understanding (TR), Prospective Reasoning (PR), Spatial Understanding (SU), Action Perception (ACP), and Counting (CT).}
\label{tab:streaming}
\small
\center
\setlength{\tabcolsep}{1.7pt}
\resizebox{\textwidth}{!}{
\begin{tabular}{lcccccccccccc}
\toprule
\textbf{Model} & \multicolumn{1}{c|}{\textbf{\#Frames}} & \textbf{OP} & \textbf{CR} & \textbf{CS} & \textbf{ATP} & \textbf{EU} & \textbf{TR} & \textbf{PR} & \textbf{SU} & \textbf{ACP} & \multicolumn{1}{c|}{\textbf{CT}} & \textbf{All} \\ \midrule
Human & \multicolumn{1}{c|}{-} & 89.47 & 92.00 & 93.60 & 91.47 & 95.65 & 92.52 & 88.00 & 88.75 & 89.74 & \multicolumn{1}{c|}{91.30} & 91.46 \\ \midrule
\multicolumn{13}{c}{\textbf{Closed-source MLLMs}} \\ \midrule
Gemini 1.5 Pro~\cite{team2024gemini15pro} & \multicolumn{1}{c|}{1} & 79.02 & 80.47 & 83.54 & 79.67 & 80.00 & 84.74 & 77.78 & 64.23 & 71.95 & \multicolumn{1}{c|}{48.70} & 75.69 \\
GPT-4o~\cite{hurst2024gpt4o} & \multicolumn{1}{c|}{64} & 77.11 & 80.47 & 83.91 & 76.47 & 70.19 & 83.80 & 66.67 & 62.19 & 69.12 & \multicolumn{1}{c|}{49.22} & 73.28 \\
Claude 3.5 Sonnet~\cite{claude3.5sonnet} & \multicolumn{1}{c|}{20} & 73.33 & 80.47 & 84.09 & 82.02 & 75.39 & 79.53 & 61.11 & 61.79 & 69.32 & \multicolumn{1}{c|}{43.09} & 72.44 \\ \midrule
\multicolumn{13}{c}{\textbf{Open-source Offline MLLMs}} \\ \midrule
% {\color[HTML]{9B9B9B} Video-LLaMA2-7B~\cite{cheng2024videollama2}} & \multicolumn{1}{c|}{{\color[HTML]{9B9B9B} 32}} & {\color[HTML]{9B9B9B} 55.86} & {\color[HTML]{9B9B9B} 55.47} & {\color[HTML]{9B9B9B} 57.41} & {\color[HTML]{9B9B9B} 58.17} & {\color[HTML]{9B9B9B} 52.80} & {\color[HTML]{9B9B9B} 43.61} & {\color[HTML]{9B9B9B} 39.81} & {\color[HTML]{9B9B9B} 42.68} & {\color[HTML]{9B9B9B} 45.61} & \multicolumn{1}{c|}{{\color[HTML]{9B9B9B} 35.23}} & {\color[HTML]{9B9B9B} 49.52} \\
% {\color[HTML]{9B9B9B} VILA-1.5-8B~\cite{lin2024vila}} & \multicolumn{1}{c|}{{\color[HTML]{9B9B9B} 14}} & {\color[HTML]{9B9B9B} 53.68} & {\color[HTML]{9B9B9B} 49.22} & {\color[HTML]{9B9B9B} 70.98} & {\color[HTML]{9B9B9B} 56.86} & {\color[HTML]{9B9B9B} 53.42} & {\color[HTML]{9B9B9B} 53.89} & {\color[HTML]{9B9B9B} 54.63} & {\color[HTML]{9B9B9B} 48.78} & {\color[HTML]{9B9B9B} 50.14} & \multicolumn{1}{c|}{{\color[HTML]{9B9B9B} 17.62}} & {\color[HTML]{9B9B9B} 52.32} \\
{\color[HTML]{9B9B9B} Video-CCAM-14B~\cite{fei2024videoccam}} & \multicolumn{1}{c|}{{\color[HTML]{9B9B9B} 96}} & {\color[HTML]{9B9B9B} 56.40} & {\color[HTML]{9B9B9B} 57.81} & {\color[HTML]{9B9B9B} 65.30} & {\color[HTML]{9B9B9B} 62.75} & {\color[HTML]{9B9B9B} 64.60} & {\color[HTML]{9B9B9B} 51.40} & {\color[HTML]{9B9B9B} 42.59} & {\color[HTML]{9B9B9B} 47.97} & {\color[HTML]{9B9B9B} 49.58} & \multicolumn{1}{c|}{{\color[HTML]{9B9B9B} 31.61}} & {\color[HTML]{9B9B9B} 53.96} \\
{\color[HTML]{9B9B9B} LongVA-7B~\cite{zhang2024longva}} & \multicolumn{1}{c|}{{\color[HTML]{9B9B9B} 128}} & {\color[HTML]{9B9B9B} 70.03} & {\color[HTML]{9B9B9B} 63.28} & {\color[HTML]{9B9B9B} 61.20} & {\color[HTML]{9B9B9B} 70.92} & {\color[HTML]{9B9B9B} 62.73} & {\color[HTML]{9B9B9B} 59.50} & {\color[HTML]{9B9B9B} 61.11} & {\color[HTML]{9B9B9B} 53.66} & {\color[HTML]{9B9B9B} 54.67} & \multicolumn{1}{c|}{{\color[HTML]{9B9B9B} 34.72}} & {\color[HTML]{9B9B9B} 59.96} \\
{\color[HTML]{9B9B9B} InternVL-V2-8B~\cite{chen2024farInternvl2}} & \multicolumn{1}{c|}{{\color[HTML]{9B9B9B} 16}} & {\color[HTML]{9B9B9B} 68.12} & {\color[HTML]{9B9B9B} 60.94} & {\color[HTML]{9B9B9B} 69.40} & {\color[HTML]{9B9B9B} 77.12} & {\color[HTML]{9B9B9B} 67.70} & {\color[HTML]{9B9B9B} 62.93} & {\color[HTML]{9B9B9B} 59.26} & {\color[HTML]{9B9B9B} 53.25} & {\color[HTML]{9B9B9B} 54.96} & \multicolumn{1}{c|}{{\color[HTML]{9B9B9B} 56.48}} & {\color[HTML]{9B9B9B} 63.72} \\
{\color[HTML]{9B9B9B} Kangaroo-V2-8B~\cite{liu2024kangaroo}} & \multicolumn{1}{c|}{{\color[HTML]{9B9B9B} 64}} & {\color[HTML]{9B9B9B} 71.12} & {\color[HTML]{9B9B9B} 84.38} & {\color[HTML]{9B9B9B} 70.66} & {\color[HTML]{9B9B9B} 73.20} & {\color[HTML]{9B9B9B} 67.08} & {\color[HTML]{9B9B9B} 61.68} & {\color[HTML]{9B9B9B} 56.48} & {\color[HTML]{9B9B9B} 55.69} & {\color[HTML]{9B9B9B} 62.04} & \multicolumn{1}{c|}{{\color[HTML]{9B9B9B} 38.86}} & {\color[HTML]{9B9B9B} 64.60} \\
{\color[HTML]{9B9B9B} LLaVA-NeXT-Video-32B~\cite{zhang2024llavanextvideo}} & \multicolumn{1}{c|}{{\color[HTML]{9B9B9B} 64}} & {\color[HTML]{9B9B9B} 78.20} & {\color[HTML]{9B9B9B} 70.31} & {\color[HTML]{9B9B9B} 73.82} & {\color[HTML]{9B9B9B} 76.80} & {\color[HTML]{9B9B9B} 63.35} & {\color[HTML]{9B9B9B} 69.78} & {\color[HTML]{9B9B9B} 57.41} & {\color[HTML]{9B9B9B} 56.10} & {\color[HTML]{9B9B9B} 64.31} & \multicolumn{1}{c|}{{\color[HTML]{9B9B9B} 38.86}} & {\color[HTML]{9B9B9B} 66.96} \\
{\color[HTML]{9B9B9B} MiniCPM-V-2.6-8B~\cite{hu2024minicpm}} & \multicolumn{1}{c|}{{\color[HTML]{9B9B9B} 32}} & {\color[HTML]{9B9B9B} 71.93} & {\color[HTML]{9B9B9B} 71.09} & {\color[HTML]{9B9B9B} 77.92} & {\color[HTML]{9B9B9B} 75.82} & {\color[HTML]{9B9B9B} 64.60} & {\color[HTML]{9B9B9B} 65.73} & {\color[HTML]{9B9B9B} 70.37} & {\color[HTML]{9B9B9B} 56.10} & {\color[HTML]{9B9B9B} 62.32} & \multicolumn{1}{c|}{{\color[HTML]{9B9B9B} 53.37}} & {\color[HTML]{9B9B9B} 67.44} \\
{\color[HTML]{9B9B9B} LLaVA-OneVision-7B~\cite{lillavaonevision}} & \multicolumn{1}{c|}{{\color[HTML]{9B9B9B} 32}} & {\color[HTML]{9B9B9B} 80.38} & {\color[HTML]{9B9B9B} 74.22} & {\color[HTML]{9B9B9B} 76.03} & {\color[HTML]{9B9B9B} 80.72} & {\color[HTML]{9B9B9B} 72.67} & {\color[HTML]{9B9B9B} 71.65} & {\color[HTML]{9B9B9B} 67.59} & {\color[HTML]{9B9B9B} 65.45} & {\color[HTML]{9B9B9B} 65.72} & \multicolumn{1}{c|}{{\color[HTML]{9B9B9B} 45.08}} & {\color[HTML]{9B9B9B} 71.12} \\
{\color[HTML]{9B9B9B} Qwen2.5-VL-7B~\cite{Qwen2.5-VL}} & \multicolumn{1}{c|}{{\color[HTML]{9B9B9B} 1}} & {\color[HTML]{9B9B9B} 78.32} & {\color[HTML]{9B9B9B} 80.47} & {\color[HTML]{9B9B9B} 78.86} & {\color[HTML]{9B9B9B} 80.45} & {\color[HTML]{9B9B9B} 76.73} & {\color[HTML]{9B9B9B} 78.50} & {\color[HTML]{9B9B9B} 79.63} & {\color[HTML]{9B9B9B} 63.41} & {\color[HTML]{9B9B9B} 66.19} & \multicolumn{1}{c|}{{\color[HTML]{9B9B9B} 53.19}} & {\color[HTML]{9B9B9B} 73.68} \\ \midrule
\multicolumn{13}{c}{\textbf{Open-source Online MLLMs}} \\ \midrule
Flash-VStream-7B~\cite{zhang2025flash} & \multicolumn{1}{c|}{-} & 25.89 & 43.57 & 24.91 & 23.87 & 27.33 & 13.08 & 18.52 & 25.20 & 23.87 & \multicolumn{1}{c|}{48.70} & 23.23 \\
VideoLLM-Online-8B~\cite{chen2024videollmonline} & \multicolumn{1}{c|}{2} & 39.07 & 40.06 & 34.49 & 31.05 & 45.96 & 32.40 & 31.48 & 34.16 & 42.49 & \multicolumn{1}{c|}{27.89} & 35.99 \\
Dispider-7B~\cite{qian2025dispider} & \multicolumn{1}{c|}{1} & 74.92 & 75.53 & 74.10 & 73.08 & 74.44 & 59.92 & 76.14 & 62.91 & 62.16 & \multicolumn{1}{c|}{45.80} & 67.63 \\
StreamAgent-7B~\cite{yang2025streamagent} & \multicolumn{1}{c|}{1} & 79.63 & 78.31 & 79.28 & 75.87 & 74.74 & 76.92 & 82.94 & 66.31 & 73.69 & \multicolumn{1}{c|}{55.40} & 74.28 \\
TimeChat-Online-7B~\cite{yao2025timechat} & \multicolumn{1}{c|}{1 
} & 80.76 & 79.69 & 80.76 & 83.33 & 74.84 & 78.82 & 78.70 & 64.23 & 68.75 & \multicolumn{1}{c|}{57.98} & {{75.28}} \\ \midrule
StreamEMS-7B (Ours) & \multicolumn{1}{c|}{1
} & 81.57 & 80.47 & 82.02 & 83.01 & 77.99 & 80.37 & 83.33 & 71.14 & 70.74 & \multicolumn{1}{c|}{45.74} & \textbf{76.20} \\ \bottomrule
\end{tabular}
}
\end{table*}

\subsection{Comparison with State-of-The-Art Methods}

\textbf{Streaming Video Understanding Benchmarks}. We evaluate our StreamEMS framework on two widely adopted streaming video understanding benchmarks: OVO-Bench and StreamingBench.

As shown in Table~\ref{tab:ovo}, our method performs better than most state-of-the-art methods in many scenarios on OVO-Bench. Compared with the basic model TimeChat-Online, our model achieves an improvement of $3.8\%$ in accuracy, i.e., from $47.6\%$  to $51.4\%$. This finding showcases the effectiveness of our method in exploiting memory capacity by mining its internal entities through self-evolving schemes. Furthermore, our method also outperforms many memory-based approaches, such as many advanced online multimodal LLM models~\cite{zhang2025flash, chen2024videollmonline, qian2025dispider}. It indicates that it is encouraging to explore more internal memory operations beyond the conventional read/write paradigm.
In addition, our method also performs better than the agentic model StreamAgent~\cite{yang2025streamagent}, achieving an accuracy of  $51.41\%$ vs. $49.4\%$.
StreamAgent employs multiple specialized agents to execute decomposed subtasks for streaming video understanding.
This finding suggests that the mechanism of internal memory refinement offers a simpler and more effective pathway to enhance streaming video understanding, rather than the development of
complex agent-based coordination.
Note that our method exhibits strong capability in handling Real-Time Visual Perception and Backward Tracing tasks, with average accuracies of $65.8\%$ and $48.21\%$, respectively.
It demonstrates that our method not only possesses an acute perceptual ability to immediately and precisely understand ongoing visual inputs but also a vast associative capacity to recall and reason about past events~\cite{niu2025ovo}. 
Consistent with our previous findings, Table~\ref{tab:streaming} also shows that our method achieves superior performance on StreamingBench, outperforming other methods on the majority of subtasks.
All these factors showcase the effectiveness of our method in video reasoning and its strong generalization ability across different datasets.

\begin{table*}[t]
\caption{Performance comparison with various models on offline video understanding benchmarks. ``↓80\%'' denotes an approximate token reduction rate of 80\%.}
\vspace{-10pt}
\label{tab:offline}
\center
\resizebox{\textwidth}{!}{
\begin{tabular}{lcccccc}
\toprule
\multicolumn{1}{l|}{\multirow{2}{*}{\textbf{Model}}} & \textbf{MVBench} & \textbf{Perception Test} & \multicolumn{2}{c}{\textbf{TempCompass}} & \textbf{LVBench} & \textbf{VideoMME} \\
\multicolumn{1}{l|}{} & Avg & Avg & Yes/No & Multi choice & Avg & Long \\ \midrule
\multicolumn{7}{l}{\textbf{Open-source Offline MLLMs}} \\ \midrule
\multicolumn{1}{l|}{LLaMA-VID-7B~\cite{li2024llamavid}} & 41.9 & 41.6 & 53.0 & 35.3 & 23.9 & - \\
\multicolumn{1}{l|}{VideoChat2-7B~\cite{li2024mvbenchVideoChat2}} & 60.4 & - & 58.0 & 51.1 & - & 33.2 \\
\multicolumn{1}{l|}{LongVA-7B~\cite{zhang2024longva}} & - & - & - & - & 35.9 & 46.2 \\
\multicolumn{1}{l|}{Kangaroo-7B~\cite{liu2024kangaroo}} & 61.1 & - & - & - & 39.4 & 46.6 \\
\multicolumn{1}{l|}{Video-CCAM-14B~\cite{fei2024videoccam}} & 64.6 & - & - & - & - & 46.7 \\
\multicolumn{1}{l|}{Video-LLaVA-7B~\cite{lin2024videollava}} & - & 44.2 & 56.4 & 44.7 & 29.3 & 36.2 \\
\multicolumn{1}{l|}{Video-ChatGPT-7B~\cite{maaz2024videochatgpt}} & 32.7 & - & 50.7 & 35.2 & - & - \\ \midrule
\multicolumn{7}{l}{\textbf{Open-source Online MLLMs}} \\ \midrule
\multicolumn{1}{l|}{MovieChat-7B~\cite{song2024moviechat}} & 55.1 & - & - & - & 22.5 & 33.4 \\
\multicolumn{1}{l|}{VideoChat-Online-4B~\cite{huang2025onlineOVBench}} & 64.9 & - & - & - & 24.0 & 44.9 \\
\multicolumn{1}{l|}{Flash-VStream-7B~\cite{zhang2025flash}} & - & - & - & - & 42.0 & - \\
\multicolumn{1}{l|}{TimeChat-Online-7B~\cite{yao2025timechat} (↓80\%)} & 60.4 & 62.8 & 66.9 & 70.9 & - & 49.2 \\
\multicolumn{1}{l|}{StreamEMS-7B (ours) (↓80\%)} & \textbf{65.2} & \textbf{66.4} & \textbf{71.6} & \textbf{76.8} & \textbf{42.2} & \textbf{51.3} \\ \bottomrule
\end{tabular}
}
\end{table*}

\textbf{Offline Video Understanding Benchmarks.}
To assess the generalizability of StreamEMS beyond streaming scenarios, we perform rigorous evaluations across a diverse range of offline video understanding benchmarks, including MVBench~\cite{li2024mvbenchVideoChat2}, Perception Test~\cite{patraucean2023perceptiontest}, TempCompass~\cite{liu2024tempcompass}, LVBench~\cite{wang2024lvbench}, and VideoMME~\cite{fu2025videomme}. 
Under these circumstances, models have full access to the entire video input.
For fair comparison, 
We assess both TimeChat-Online and StreamEMS under comparable token budgets, corresponding to an approximate token reduction rate of $80\%$.
As shown in Table~\ref{tab:offline}, StreamEMS consistently achieves strong performance across different datasets. 
For example, StreamEMS achieves an accuracy of 
$65.2\%$ on MVBench, surpassing TimeChat-Online by a margin of  $4.8\%$
and also outperforming the strong offline baseline, Video-CCAM-14B.
The superiority of our method is also demonstrated in scenarios requiring fine-grained temporal reasoning.
On the Multi-choice split of TempCompass, our method achieves a gain of 
$5.9\%$, increasing from  $70.9\%$ to $76.8\%$, while on the long-form VideoMME benchmark, StreamEMS achieves 
$51.3\%$, reaching a $2.1\%$ improvement over TimeChat-Online.
Furthermore, our model also demonstrates advanced capabilities in the domain of long video understanding.
Specifically, our model achieves $42.2\%$ on LVBench, surpassing Flash-VStream as well as several offline counterparts, including Video-LLaVA and LongVA.
Collectively, these results validate the effectiveness of our method in generalizing across online and offline video understanding scenarios.

\subsection{Ablation Study}
\label{sec:exp.ablations}
In this section, we present extensive ablation studies to evaluate the effectiveness of our proposed designs.  Unless stated otherwise, we perform the ablation studies on OVO-Bench. We adopt the competitive model TimeChat-Online as our baseline model.

\begin{table}[t]
  \caption{Investigation on the effectiveness of the self-evolving schemes. SEM denotes Semantic Evolution Module, while PEM refers to Prior-informed Evolution Module.}
  \label{tab:investigation_on_proposed_strategies}
  \centering
  \resizebox{0.5\linewidth}{!}{
  \begin{tabular}{@{}lc@{}}
    \toprule
    \textbf{Method}                 & \textbf{Accuracy}  \\
    \midrule
   Our Baseline                    & 48.46                                   \\
    Our Baseline + PEM                & 50.56                                 \\
    Our Baseline + SEM                & 51.22                                 \\
    Our Baseline + PEM + SEM & 51.41                 \\
    \bottomrule
  \end{tabular}
  }
\end{table}

\textbf{Effectiveness of Self-evolving Schemes.} We propose to examine the effectiveness of our proposed self-evolving schemes, which are implemented by the Semantic Evolution Module and Prior-informed Module.
 As shown in Table~\ref{tab:investigation_on_proposed_strategies}, the baseline model achieves an accuracy of $48.46\%$. On top of that, we gradually employ the proposed modules. As expected, equipping the baseline with SEM and PEM individually brings consistent improvements, yielding $2.76\%$ and $2.10\%$ performance gains, respectively. This factor clearly indicates the effectiveness of exploiting the intrinsic capacity of memory through our proposed designs. Moreover, the two modules complement each other, achieving a $2.95\%$ improvement when integrated. It suggests that boosting native memory should not only rely on propagating semantic information across the semantic graph to enhance information density, but also on incorporating the temporal priors preserved in previous memory states.

\begin{table}[t]
  \caption{Investigation on the hierarchy of semantic threshold $\theta$. We propose to perform progressive semantic enhancement through the hierarchical selection of $\theta$ at different semantic scales. The threshold values are determined by analyzing the similarity distribution and selecting the corresponding values at specific quantile points.}
  \label{tab:investigation_on_semantic_threshold}
  \centering
  \resizebox{0.45\linewidth}{!}{
  \begin{tabular}{@{}lc@{}}
    \toprule
   \textbf{Semantic Threshold} $\theta$               & \textbf{Accuracy} \\
    \midrule
    0.85 & 51.17 \\
    (0.85, 0.9) & 51.19 \\
    (0.85, 0.9, 0.96) & \textbf{51.41} \\
    (0.85, 0.9, 0.96, 0.98) & 51.38 \\
    \bottomrule
  \end{tabular}
  }
\end{table}

\textbf{Investigation on the Hierarchy of Semantic Thresholds.} In the Semantic Evolution Module, we utilize the semantic threshold $\theta$ to 
identify
highly relevant
memory entities for semantic enhancement. 
In addition, we achieve progressive semantic enhancement by dynamically adjusting $\theta$ through a hierarchical selection mechanism at varying semantic granularities. We determine these threshold values by statistically analyzing the similarity distribution and selecting them according to specific quantile points of that distribution. We find that the similarity between memory entities will become relatively high at the end of the training phase. In this case, we empirically set the basic semantic threshold as $0.85$, and gradually increase the value. As shown in Table~\ref{tab:investigation_on_semantic_threshold}, the accuracy increases when introducing the semantic threshold in a progressive and hierarchical manner. The accuracy reaches its maximum with the semantic threshold configuration of $(0.85, 0.90, 0.96)$. The hierarchical semantic thresholds allow the memory to evolve by encapsulating useful and distinct information from existing memory entities, resulting in a more compact and information-dense representation. However, the accuracy slightly drops when using the configuration with higher semantic thresholds. This performance drop is likely attributable to the overly strict selection criterion caused by a higher semantic threshold, which may discard potentially useful memory entities and consequently compromise the overall memory representation in turn. Therefore, it is essential to take reasonable semantic granularities into consideration when designing the semantic hierarchy.

\begin{figure}[t]
    \centering
    \begin{minipage}[t]{0.48\textwidth}
        \centering
        \includegraphics[width=\textwidth]{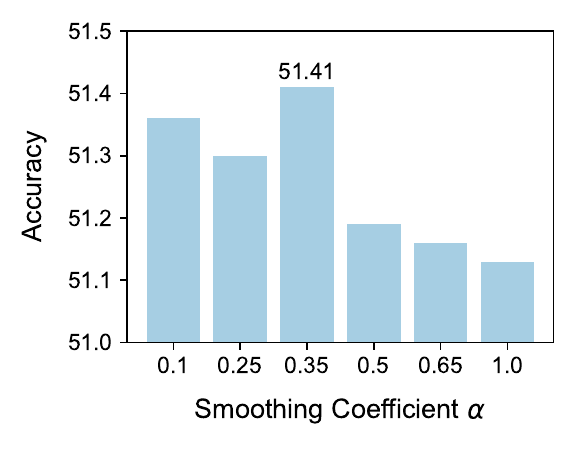}
        \vspace{-0.3cm}
        \caption{Investigation on the smoothing coefficient $\alpha$ in the Prior-informed Evolution Module.}
        \label{fig:ablation-pem-alpha}
        \vspace{-0.32cm}
    \end{minipage}\hfill\hfill
    \begin{minipage}[t]{0.48\textwidth}
        \centering
        \includegraphics[width=\textwidth]{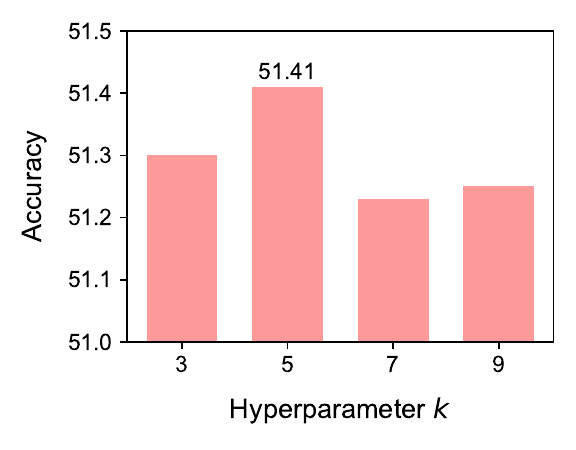}
        \vspace{-0.35cm}
        \caption{Investigation on the hyperparameter $K$ in Piror-informed Evolution Module.}
        \label{fig:ablation-pem-k}
        \vspace{-0.3cm}
    \end{minipage}
\end{figure}

\textbf{Investigation on the Smoothing Coefficient $\alpha$ in PEM}. As discussed, we use the smoothing coefficient $\alpha$ to integrate the historical similarity distribution to refine the current memory state. In this case, we propose to examine its effectiveness on OVO-Bench.
According to Eq.~\ref{eq:ema}, when $\alpha$ is set to $1$, the scenario degenerates into the conventional case, where only the current information is used without taking into account previous memory distribution.
As shown in Fig.~\ref{fig:ablation-pem-alpha}, this setting performs the worst. It indicates that it is essential to leverage prior knowledge. When $\alpha$ decreases progressively, a greater amount of historical information is incorporated. 
We observe that $\alpha = 0.35$ achieves the best performance.
Therefore, we empirically adopt this value in our work.

\textbf{Investigation of the Hyperparameter $K$ in PEM.}
After applying the EMA strategy to update the current memory state, we further employ a Top-K strategy to select relatively distinct memory entities for interaction with the new input. As shown in Fig.~\ref{fig:ablation-pem-k}, we vary the number of $K$ to examine its influence. The performances under different configurations are relatively similar, with the setting of $K=5$ achieving the best result.
Therefore, we empirically adopt this value in our work.

\begin{figure}[t]
    \centering
    \begin{minipage}[t]{0.48\textwidth}
        \centering
        \includegraphics[width=\linewidth]{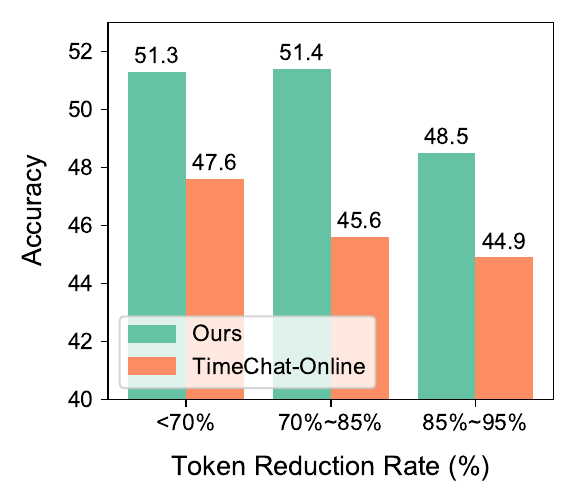}
        \vspace{-0.2cm}
        \caption{Comparisons across different token reduction rate settings on OVO-Bench}
        \label{fig:ablation-tkdrop-ovo}
    \end{minipage}\hfill
    \begin{minipage}[t]{0.48\textwidth}
        \centering
        \includegraphics[width=\linewidth]{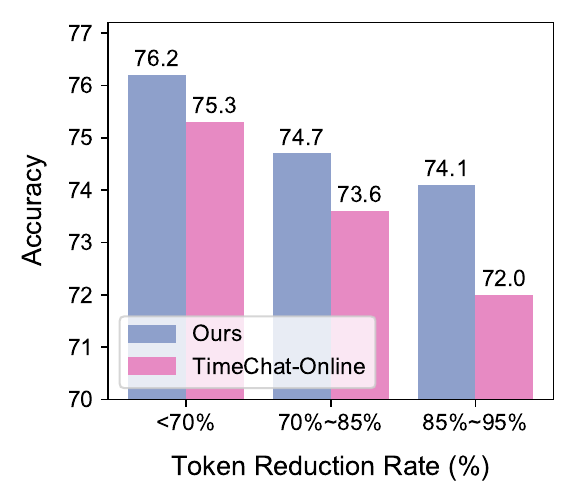}
        \caption{Comparisons across different token reduction rate settings on StreamingBench}
        \label{fig:ablation-tkdrop-str}
    \end{minipage}
\end{figure}

\textbf{Investigation on the Robustness.}
The token budget serves as a crucial indicator for evaluating the model's real-time performance in streaming video understanding scenarios. Accordingly, many methods focus on reducing the number of tokens while preserving overall performance. The token reduction plot evaluates robustness by illustrating how performance is preserved relative to the reduction of input tokens. Therefore, we assess the robustness of our method by integrating it with TimeChat-Online and evaluating performance across OVO-Bench and StreamingBench under varying token drop rates.
As shown in Fig.~\ref{fig:ablation-tkdrop-ovo} and Fig.~\ref{fig:ablation-tkdrop-str}, it is clear that our method consistently outperforms TimeChat-Online in all cases, which reflects the strong robustness of our method regardless of the dropping rate.
In addition, we observe that the performance gap widens significantly under aggressive token reduction. Specifically, our method outperforms the baseline by $5.81\%$ at $70\%$--$85\%$ dropping rates on OVO-Bench, and maintains a $2.12\%$ margin even at extreme rates of $85\%$--$95\%$. This observation suggests that our method achieves higher token efficiency under limited-token settings, leading to improved performance. 
This is mainly because our method makes full use of memory entities and evolves the memory into a more informative representation through the proposed designs.

\begin{figure}[t]
    \centering
    \includegraphics[height=0.4\linewidth]{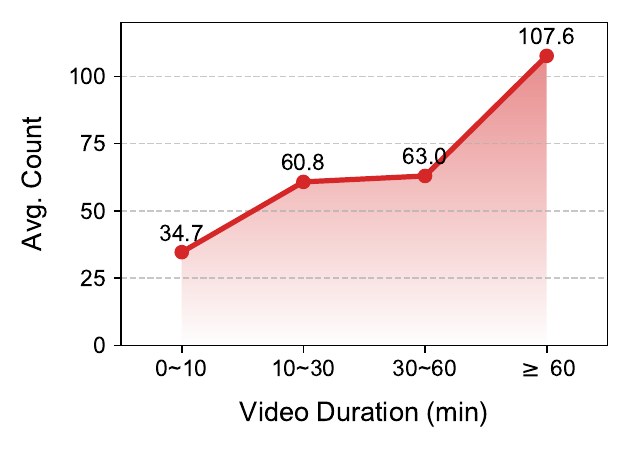}
    \vspace{-0.3cm}
    \caption{The average count of memory entities of our method across different video durations. The metrics are obtained under the same setting as Table~\ref{tab:efficiency}.}
    \label{fig:ablation-memcount}
    \vspace{-0.25cm}
\end{figure}

\begin{table}[t]
\caption{Performance comparison of our method and the baseline (TimeChat-Online, TCO) on MLVU videos of varying durations. All metrics are averaged over 20 randomly sampled videos per duration bin (80 videos total). ↓ indicates lower is better, ↑ indicates higher is better.
}
\centering
\resizebox{0.8\linewidth}{!}{
\label{tab:efficiency}
\begin{tabular}{l|c|cccc}
\toprule
\multirow{2}{*}{\textbf{Item}} & \multirow{2}{*}{\textbf{Method}} & \multicolumn{4}{c}{\textbf{Video Duration}} \\ \cline{3-6} 
 &  & \textbf{0$\sim$10 min} & \textbf{10$\sim$30 min} & \textbf{30$\sim$60 min} & \textbf{\textgreater{}= 60 min} \\ \midrule
\multirow{2}{*}{\textbf{\begin{tabular}[c]{@{}l@{}}Multimodal Token\\ VRAM (MB) (↓)\end{tabular}}} & TCO~\cite{yao2025timechat} & 82.7 & 132.1 & 133.4 & 188.8 \\
 & Ours & \textbf{27.1} & \textbf{38.3} & \textbf{36.5} & \textbf{57.9} \\ \midrule
\multirow{2}{*}{\textbf{\begin{tabular}[c]{@{}l@{}}Token Reduction\\ Rate (\%) (↑)\end{tabular}}} & TCO~\cite{yao2025timechat} & 88.3\% & 82.0\% & 81.4\% & 74.1\% \\
 & Ours & \textbf{95.7\%} & \textbf{94.8\%} & \textbf{94.9\%} & \textbf{92.7\%} \\ \midrule
\multirow{2}{*}{\textbf{Average FPS (↑)}} & TCO~\cite{yao2025timechat} & 23.0 & 24.2 & \textbf{27.2} & 25.5 \\
 & Ours & \textbf{34.5} & \textbf{34.0} & 28.3 & \textbf{24.7} \\ \midrule
\multirow{2}{*}{\textbf{TTFT (s) (↓)}} & TCO~\cite{yao2025timechat} & 2.2 & 3.7 & 3.8 & 5.7 \\
 & Ours & \textbf{1.4} & \textbf{1.7} & \textbf{1.8} & \textbf{2.7} \\ \midrule
\multirow{2}{*}{\textbf{\begin{tabular}[c]{@{}l@{}}Token Generation\\ Interval (ms) (↓)\end{tabular}}} & TCO~\cite{yao2025timechat} & 166.7 & 202.0 & 161.6 & 142.3 \\
 & Ours & \textbf{64.3} & \textbf{64.6} & \textbf{58.8} & \textbf{62.1} \\ \bottomrule
\end{tabular}
}
\end{table}

\textbf{Investigation on Memory Consumption and Inference Latency.} To evaluate memory footprint and inference latency, we compare our method against the streaming baseline, TimeChat-Online (TCO). All experiments are conducted on a single NVIDIA A6000 GPU (50GB) using video samples from the MLVU benchmark. To examine the impact of video length, we partition the dataset into four duration-based bins: 0--10, 10--30, 30--60, and $\ge$60 minutes. We randomly sampled 20 videos from each bin, yielding a total of 80 test sequences. Additionally, we adjust the token drop threshold of TCO to 0.25 (from the default 0.5) to achieve an approximate 80\% reduction rate. This ensures a more comparable efficiency against our method, which maintains a higher reduction rate of around 90\%. Detailed results can be found in Table~\ref{tab:efficiency} and Fig.~\ref{fig:ablation-memcount}.

First, to evaluate our memory mechanism, we analyze the average number of generated memory entities across varying video durations. As shown in Fig.~\ref{fig:ablation-memcount}, while the entity count scales with video length (ranging from 34.7 to 107.6), the growth remains controlled. This demonstrates that our method avoids unbounded expansion regardless of the video duration. Second, as shown in Table~\ref{tab:efficiency}, our method exhibits consistent memory efficiency in terms of multimodal token consumption across varying video durations, e.g., incurring nearly one-third the memory cost of the baseline.
Meanwhile, the reduction rate of our method is nearly invariant to video length ($95.68\%$ to $92.66\%$); whereas the counterpart for TCO drops significantly, falling from $88.3\%$ to $74.1\%$ as the duration scales. 
These memory efficiency gains stem from our clip-level representation strategy for capturing long-range historical context, rather than relying on the traditional memory-intensive frame-by-frame processing paradigm.
Third, our approach exhibits superior inference throughput.
To be specific, the average FPS of our method is consistently larger than that of the baseline model in most cases. In addition, our approach maintains a consistently low TTFT irrespective of video duration, achieving an approximate $2\times$ reduction in latency. Time-to-First-Token (TTFT) serves as a critical metric for quantifying the real-time responsiveness of a model. Finally, we evaluate the token generation interval during LLM decoding. Our method consistently achieves lower inter-token latency compared to TCO, demonstrating its superior per-token generation speed.

\begin{figure*}[t]
    \centering
    \includegraphics[width=\linewidth]{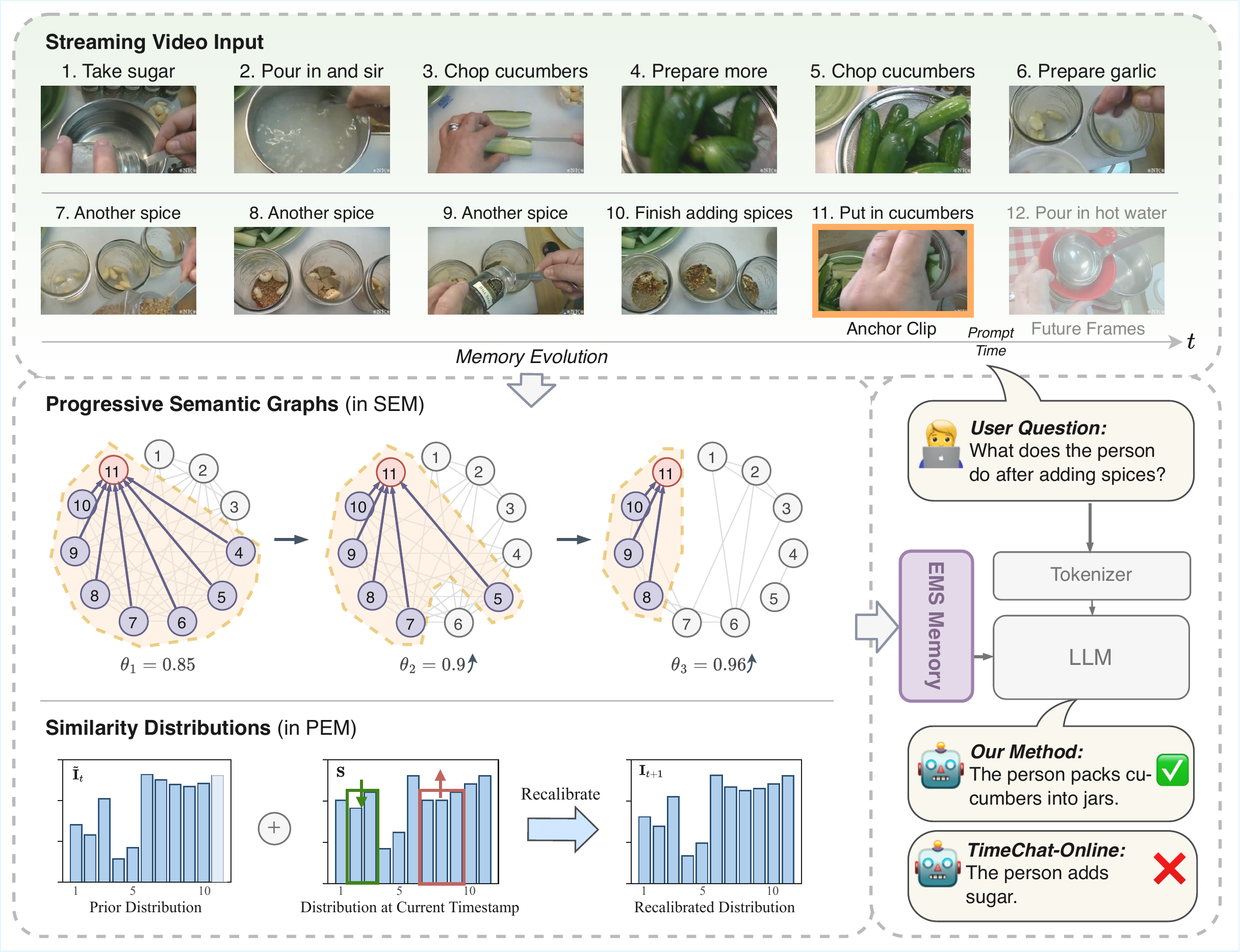}
    \caption{\textbf{Visualization of a successful case}. Compared with the baseline model TimeChat-Online, our method makes the correct prediction by focusing on highly semantically related scenarios with the assistance of self-evolving memory schemes. On the one hand, the progressive semantic evolution strategy enables the model to gradually aggregate distinct information from highly similar memory entities. On the other hand, the prior-informed evolution strategy allows the model to recalibrate the original similarity distribution using the previous memory distribution.}
    \label{fig:visualization}
\end{figure*}

\section{Visualization}

To verify the effectiveness of our method, we provide a qualitative result to illustrate how our designs work. As shown in Fig.~\ref{fig:visualization}, the question inquires what the person does after adding the spices.
The baseline model TimeChat-Online gives an incorrect answer because it may focus on irrelevant actions, like adding sugar. In contrast, our model derives the correct answer by evolving the memory through the proposed schemes. First, our method uses progressive evolution strategies to gradually aggregate distinct information
from semantically related clips. The progressive evolution is achieved by using a sequence of stricter thresholds for information aggregation,
making the memory representation more compact and informative. 
In Fig.~\ref{fig:visualization}, we illustrate the integration process when the 11th clip is produced, where we denote it as the Anchor Clip in the figure. Focusing on the 11th clip for clarity, we observe that as the semantic thresholds progress, its connections increasingly concentrate on semantically relevant clips, 
thereby iteratively distilling discriminative information from these neighboring clips into the 11th clip.
In addition, the original similarity distribution exhibits a temporal bias,
mistakenly assigning lower weights to temporally adjacent clips while favoring temporally distant clips.
This reliance on the current distribution may cause the model to attend to irrelevant clips, such as adding sugar. Instead, our method leverages the priors encoded in the previous distribution in order to recalibrate the current distribution, aligning it more closely with the true temporal context. The proposed self-evolving memory schemes facilitate a more discriminative representation, ultimately leading to the correct prediction and validating the efficacy of our approach.

\section{Conclusion}

In this work, we present StreamEMS, a general mechanism for improving streaming video understanding by re-structuring the historical data stored in memory through self-evolving memory scheme, enabling more informative and robust memory representations.
Specifically,
We first introduce a Semantic Evolution Module to evolve the memory into more information-dense representations. It works by exploiting informative memory entities using hierarchical semantic graphs, which are constructed with a coarse-to-fine hierarchy of semantic thresholds.
In addition, we further introduce a Prior-informed Evolution Module to evolve memory into more robust representations by leveraging prior memory distributions to refine the current memory state.
We validate the effectiveness of our proposed designs on widely-used streaming video understanding datasets. As expected, our method outperforms other methods, showcasing the effectiveness of our designs.





\section{Declaration of Generative AI and AI-Assisted Technologies in the Manuscript Preparation Process}

During the preparation of this work, the authors used Large Language Models (LLMs) solely for general-purposed grammar correction, language refinement and stylistic improvements during the writing process. The authors reviewed and edited the output as needed and take full responsibility for the content of the published article.

\bibliographystyle{elsarticle-num}
% \bibliography{reference} 

\end{document}